# Incorporating characteristics of human creativity into an evolutionary art algorithm

Steve DiPaola
Simon Fraser University, CANADA
www.sfu.ca/~sdipaola
sdipaola@sfu.ca

Liane Gabora
University of British Columbia, CANADA
https://people.ok.ubc.ca/lgabora/
liane.gabora@ubc.ca

**Abstract**
A perceived limitation of evolutionary art and design algorithms is that they rely on human intervention; the artist selects the most aesthetically pleasing variants of one generation to produce the next. This paper discusses how computer generated art and design can become more creatively human-like with respect to both process and outcome. As an example of a step in this direction, we present an algorithm that overcomes the above limitation by employing an automatic fitness function. The goal is to evolve abstract portraits of Darwin, using our 2nd generation fitness function which rewards genomes that not just produce a likeness of Darwin but exhibit certain strategies characteristic of human artists. We note that in human creativity, change is less choosing amongst randomly generated variants and more capitalizing on the associative structure of a conceptual network to hone in on a vision. We discuss how to achieve this fluidity algorithmically.

## 1 Introduction

Through neuroimaging studies, theoretical and empirical work on the psychology of creativity, and simulation, we are gaining insight into the once mysterious process of creativity. Human creativity results in items that are not just unusual, but useful or aesthetically pleasing. To what extent can this be mimicked by a computer? To what extent is human input necessary to make decisions along the way as to how the computer program should proceed next? In this paper we describe a portrait painter evolutionary algorithm that does not rely on human intervention. We also discuss further steps for generating complex and aesthetically appealing computer art using algorithms that employ human-like creative mechanisms. The paper concludes with a discussion of how research on the mechanisms underlying human creativity and the personality traits of creative individuals can inform the development of computational systems for art and design generation.

## 2 Creativity research

The application of genetic programming and/or evolutionary approaches to the generation of art would appear to be most compatible with the view that creative thought is a Darwinian process [29–31]. According to this view, we generate new ideas through Blind1 Variation and Selective Retention (thus it is abbreviated BVSR): i.e. randomly alter the current thought a multitude of different ways, select the fittest variant(s), and repeat this process until a satisfactory idea results. The work described here builds upon previous efforts that were closely aligned with this kind of Darwinian approach, but departs from it using insights from other approaches to creativity.

It has long been believed that creativity involves an uncensored, ‘primary process’, or associative form of thought, in which, as famously described by mathematician Henri Poincare: ‘‘ideas rise from the crowds’’, and seem to float, collide and connect. This is supported by neuroimaging technologies such as

Reference:
DiPaola, S. & Gabora, L. (2009). Incorporating characteristics of human creativity into an evolutionary art algorithm. Genetic Programming and Evolvable Machines, 10(2), 97-110.
fMRI studies, which permit neuroscientists to visualize human brain activity in awake individuals as they conduct various tasks. These studies show that associative thought occurs when multiple regions of the highly developed human association cortex interact with one another [1]. Andreasen proposes that at this point the brain is functioning as a self-organizing system at the proverbial 'edge of chaos' [18], where associative thought processes reign. At the neural level, associations begin to form where they did not previously exist, allowing for enriched connectivity between the various association cortices, or even creating new kinds of connectivity. An interesting finding that arises from case-study research is that creators often work within a very structured domain, following rules that they eventually break free of [7, 9]. They use the template of a sonnet, opera, symphony, comedy or tragedy as a base from which to depart and elaborate. For example, when Michelangelo created his sculpture of David, he took as a departure point, the archetypal narrative as well as sculptures of David by other great Renaissance masters. His genius lay in that he was able to break loose from an established form to create something new with respect to both in conceptualization and execution. Thus creativity is not just a matter of eliminating rules but of assimilating and then breaking free of them where warranted. Indeed a considerable body of research suggests that the creative process involves not just increased fluidity or free associative thought, but increased fluidity tempered with increased restraint. As Feist [8] puts it: ''It is not unbridled psychoticism that is most strongly associated with creativity, but psychoticism tempered by high ego strength or ego control. Paradoxically, creative people appear to be simultaneously very labile and mutable and yet can be rather controlled and stable''. The existence of two stages of the creative process is consistent with the widely held view that there are two distinct forms of thought [5, 24–26, 33]. It has been proposed that creativity involves the ability to vary the degree of conceptual fluidity in response to the demands of any given phase of the creative process [9, 10, 12, 13]. This is referred to as contextual focus [10, 12, 13]. Focused attention produces analytic thought, which is conducive to manipulating symbolic primitives and deducing laws of cause and effect, while defocusing attention produces fluid or associative thought which is conducive to analogy and unearthing relationships of correlation. It is this conceptual fluidity of contextual focus that we have implemented within our 2nd iteration of the abstract portrait fitness function algorithm.

The theory that one can shift between associative and analytic thought depending on the situation is supported by the neuroimaging studies mentioned previously, which show that the brain is constantly adapting and changing in response to the demands and pressures of the environment that it encounters. It was demonstrated that activated cell assemblies are composed of multiple 'neural cliques', groups of neurons that respond differentially to general or context-specific aspects of a situation [19, 20]. Neural cliques that would not be included in the assembly if one were in an analytic mode, but would be if one were in an associative mode, are referred to as neurds [12]. It is posited that the shift to a more associative mode of thought conducive to insight is accomplished by recruiting neurds that respond to abstract or atypical subsymbolic microfeatures of the problem or situation. Since memory is distributed and content-addressable this fosters reminding and the forging of creative connections to potentially relevant items previously encoded in those neurons. Thus it is tentatively proposed that creative thought involves neither randomness [29–31], nor search through a space of predefined alternatives [28, 38– 40]), but emerges naturally through the recruitment of neurds. It is suggested that this occurs when there is a need to resolve conceptual gaps in ones' internal model of the world, and resolution involves context-driven actualization of the potentiality afforded by its fine-grained associative structure.

## 3 Computer art research

The algorithm described here falls under the class of creative evolutionary systems [4]. Creative evolutionary systems grew out of evolutionary computation, a class of search algorithms inspired by Darwinian evolution, the most popular of which are genetic algorithms (GA) and genetic programming (GP) [16]). These techniques solve complex problems by encoding a population of randomly generated potential solutions as 'genetic instruction sets', assessing the ability of each to solve the problem using a predefined fitness function, mutating and/or marrying (applying crossover to) the best to yield a new generation, and repeating until one of the offspring yields an acceptable solution.

Whereas GAs and GP are generally employed to solve optimization problems, creative evolutionary systems are used to evolve aesthetically pleasing or innovative structures. In order to favor innovative solutions, rather than building in a preconceived notion of what is optimal, they lean toward starting with low-level building blocks for constructing solutions, and tend toward a relaxation or removal of constraints, both of which facilitate exploration. While strong constraining or parameterization of a function set allows for fast and optimized results, it limits the available search space and hence the ability of the system to come up with solutions that are 'outside the box'. The hazard of this approach is a very large search space (creating longer runs) or worse, the possibility of getting caught in local minima. (We will discuss shortly how this is dealt with.)

Although creative evolutionary systems are unquestionably useful as tools to enhance our own creative processes [4], and have generated some impressive art, music, and design [27, 32, 34], the extent to which they are genuinely creative is a source of debate [21]. In artistic domains, one faces a challenge that does not arise with optimization problems [23], such as the question of how to write a logical fitness function that has an aesthetic sense. To get around this problem, unlike other forms of evolutionary computation, they require a human to guide the direction of the evolutionary search [3, 15, 32]). While this kind of computer/human collaboration is an interesting and successful technique, it has several disadvantages. A first is speed; the system stops at every run and waits for a human to judge the results. A second is coverage: it is impossible to give a human all the possibilities to judge from, so most systems of this type limit the population (i.e. 8–16 individuals).A third problem is that in this context humans tend to base decisions on how things appear at that moment rather than on long term potential to evolve. A final (and to us, fundamental) shortcoming of the human 'creative decision maker' approach is that it sidesteps the goal of understanding the creative process to the point where we can make a computer be genuinely creative.

## 4 Building aesthetic sensibility into a computer art program

The portrait painter system described here uses an automatic fitness function, albeit specific to portrait painting, thereby addressing this perceived shortcoming of creative evolutionary systems and directly exploring to what extent computer algorithms can be creative on their own (Fig. 1). Others have begun to use creative evolutionary systems with an automatic fitness function in design and music [4], as well as building of a creative invention machine [17]. What is unique in our approach is that it incorporates techniques inspired by both creativity research, and capitalizes on recent developments in GP by employing a form of GP called Cartesian Genetic Programming (CGP) [22, 36; see also 37]. Genetic programming starts with a population of randomly generated computer programs composed of the available ingredients from the created function set. Genetic programming iteratively transforms a population of programs into a new generation by applying Darwinian techniques.

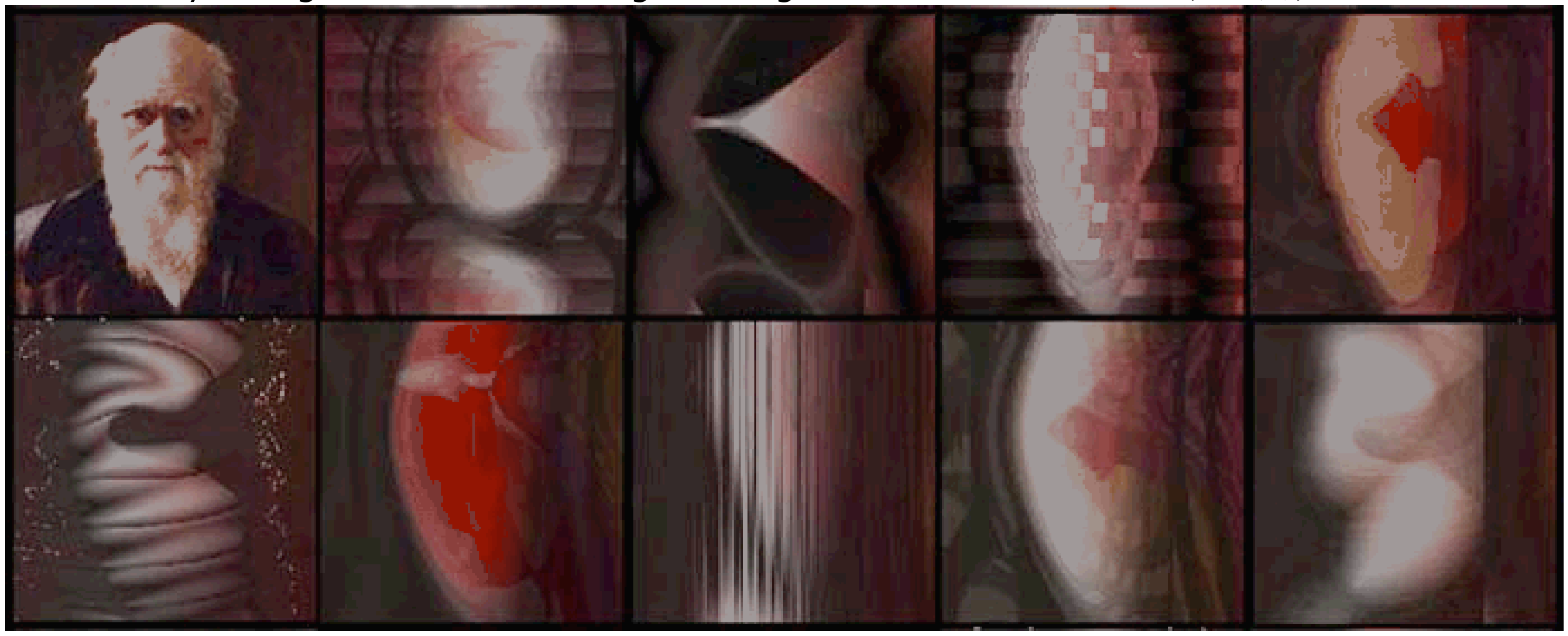

Fig. 1 Source Darwin image with examples of evolved abstract portraits created using an automatic creative system

CGP uses GP techniques (crossover, mutation, and survival), but differs in certain key respects. The program is represented by a directed graph of indexed nodes. Each node has a number of inputs and a function that gives an output based on the inputs. The genotype is a list of integers that determine the connectivity and functionality of the nodes, which can be mutated and mated to create new directed graphs. In our implementation crossover occurs between whole nodes and the mutation rate is controlled by adaptive mutation. CGP has several features that foster creativity including (1) because of its node based structure it facilitates the creation of visual mapping modules, (2) its structure accommodates a sophisticated color space model which enables painterly decision making, and most importantly (3) its component based approach favors exploration over optimization by allowing different genotypes to map to the same phenotype. The last technique uses redundancy at the input, node, and functional levels, allowing the genotype to contain nodes that are not connected to the output nodes and so not expressed in the phenotype. Having different genotypes (recipes) map to the same phenotype (output) provides CGP with greater neutrality [22, 35, 41, 42]. The advantage is that when a plateau or local minima is reached, unexpressed nodes in the genotype change, potentially leading to an improvement in fitness later on when expressed, and a chance to escape local minima. In our 2nd iteration fitness function (described below) we are able to move towards more associative ‘fuzzy rules of art’ fitness testing over the more focused resemblance function testing based on these redundancy triggers.

Our work is based on Ashmore and Miller’s [2] CGP application to evolve visual algorithms for enhanced image complexity or circular objects in an image. Ashmore and Miller’s function set, we incorporated, which in a slightly reduced form into our system, uses CGP graphs that have two inputs: the x and y coordinates of a pixel in the image, and three outputs: the hue, saturation and value (H, S, V) color channels for that pixel. An integer array is used to store the genotype with a length (n * 4) ? 3, where n is the number of nodes and the last three integers are pointers to the output HSV color values. The function set has 13 functions which use unitized x and y positions of the output image as variables and additional parameter variables (noted param) that can be affected by adaptive mutation. Functions are specifically low level in nature which aids in a large ‘creative’ search space and output HSV values between 0 and 255. An individual in our population is manifested as one program that runs successively for every pixel in the output image, which is then tested against our fitness function. This allows correlated painterly effects as you move through the image. Functions 1 through 5 use simple logical or arithmetic manipulations of the positions (low level functions create a larger ‘creative’ search space), whereas 7 through 14 use trigonometric or logical functions that are more related to geometric shapes and color graduations. The 13 functions of the function set are:

1: x|y;

```
2: param & x;
3: (x ? y) % 255;
4: if (x[y) x - y; else y - x;
5: 255 - x;
6: abs (cos (x) * 255);
7: abs (tan (((x % 45) * pi)/180.0) * 255));
8: abs (tan (x) * 255) % 255);
9: sqrt ((x - param)2 ? (y - param) 2); (thresholded at 255)
10: x % (param ? 1) ? (255 - param);
11: (x ? y)/2;
12: if (x[y) 255 * ((y ? 1)/(x ? 1)); else 255 * ((x ? 1)/(y ? 1));
13: abs (sqrt (x - param2 ? y - param2) % 255);
```

Most of Ashmore and Miller's efforts involve initializing a population and then letting the user take over. Our initial prototype was based upon their approach, but expanded it with a more sophisticated similarity and creativity function, and revised their system for a portrait painter process. Our second iteration, which is the focus of this paper, introduces a fitness function algorithm that incorporates the findings from creativity research discussed earlier, including contextual focus (a variable level of fluidity and control over different phases of the creative process). The fitness function varies fluidly from tightly focusing on resemblance (similarity to the sitter image, which in this case is the Darwin portrait), to swinging (based on functional triggers) toward a more associative process, intertwining, and at times contradicting, 'rules' of abstract portrait painting (details below). Different genotypes map to the same phenotype allows us to vary the degree of creative fluidity because it offers the capacity to move though the search space via genotype (small ordered movement) or phenotype (large movement but still related). For example, in one set of experiments this is implemented as follows: if the fittest individual of a population is identical to an individual in the previous generation for more than three iterations, other genotypes that map to this same phenotype are chosen over the current non-progressing genotype. The next section discusses other areas that we incorporated from research on creativity.

## 5 Implementation

The automatic fitness function partly uses a 'portrait to sitter' resemblance. Our fitness function must give a specific and correlated score at any resolution level to be effective; judging painterly similarity of any portrait image in deciding which individuals are more fit even in very early runs. This non-trivial assessment at early runs is one of the reasons why creative fitness functions are still very difficult to write. Among other open ended problems [21], creative fitness functions must judge even very early arbitrary results with full accuracy.

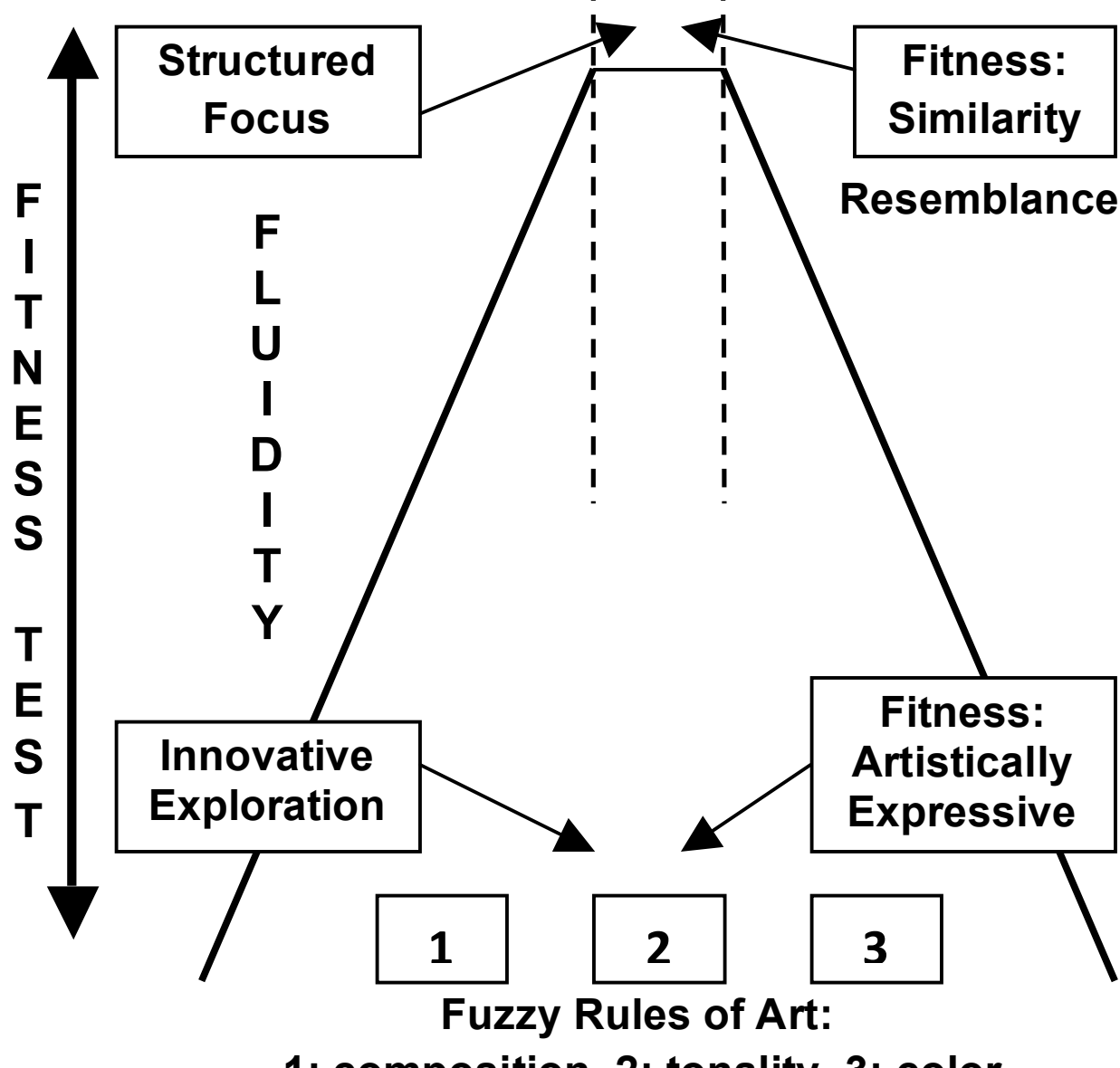

Fig. 2 Our 2nd generation fitness function mimics human creativity by moving between restrained focus (resemblance) to more unstructured associative focus (resemblance + more ambiguous art rules of composition, tonality and color theory).

theory)Since the advent of photography, portrait painting has not just been about accurate reproduction, but also about using modern painterly goals to achieve a creative representation of the sitter. We have created a fitness function that mainly rewards accurate representation, but given certain situations, also rewards visual painterly aesthetics using simple rules of art creation as well as a portrait knowledge space. Specifically, the painterly portion of our fitness function: (1) takes account of face versus background composition, (2) tonal similarity over exact color similarity, matched with a sophisticated artistic color space model which weighs for warm-cool color temperature relationships based on analogous and complementary color harmony rules, and (3) unequal dominate and subdominant tone and color rules and other artistic rules based on a portrait painter knowledge domain as detailed in [6] and illustrated in Fig. 2. It is also possible (and possibly the direction of our next version of the program) to evolve these creative evaluations simultaneously with the system. This can alter the dimensionality of a space, the parameterization, as well as the representation of solutions, allowing for more creative automation. However, in this current version, we are biased heavily toward resemblance, which gives us a structured system, but can under the influence of functional triggers allow for artistic creativity. The approach gives us novelty and innovation from within, or better said, responding to a structured system—a trait of human creative individuals. Portrait programs in the beginning of the run will look less like the sitter but from an aesthetic point of view might be highly desirable since the function set has been built with painterly rules.

Specifically, the fitness function calculates four scores (resemblance and the three painterly rules) separately and fluidly combines them in different ways to mimic human creativity by moving between restrained focus (resemblance) to more unstructured associative focus (resemblance ? rules of composition, tonality and color theory). In its default state the fitness function uses a ratio of 80% resemblance to 20% non-proportional scoring of our three painterly rules. Several functional triggers can alter this ratio in different ways. Even within any run, for instance as long as an adaptive percentage of 80–20 resemblance bias is maintained (resemblance patriarchs), the system will allow very high scoring of painterly rule individuals (strange uncles) to be accepted into the next population. These individual with high painterly scores (weighted non-proportionally including for a very high score in just one rule) are saved separately, and mated with the current 80/ 20 population; unless other triggers exist their offspring are still tested with the 80– 20 resemblance test. The main system wide functional changes occur when redundancy triggers affect the default ratio for all individuals. As mentioned in the previous section, when a plateau or local minima is reached for a certain number of populations, the fitness function ratio switches course where painterly rules are weighted higher than resemblance (on a sliding scale) and work in conjunction with redundancy at the input, node, and functional levels. Similarly, but now in reverse, to the default resemblance situation, very high scoring resemblance individuals within any run are allowed to pass into the next population when a percentage of painterly rules individuals is met. Using this method, in this more wide associative mode, high resemblance individuals are always part of the mix and when these individuals show a marked improvement in resemblance, a trigger is set to return to the more focused 80/20 resemblance ratio.

Our first prototype ran on one high-end PC for 50 days. Our 2nd iteration as described above ran for approximately the same duration. Since the genes of each portrait can be saved, it is possible to re-combine (marry) and re-evolve any of the art works in new variants (Fig. 3). As the fitness score increases, portraits look more like the sitter (Fig. 4). This gives us a somewhat known spread from very primitive (abstract) all the way through realistic portraits. So in effect our system has two ongoing progressing processes: (1) those 'most fit' portraits that pass on their portrait resemblance strategies, making for more and more realistic portraits—the family 'resemblance' patriarchs (Fig. 4), and (2) the creative 'strange uncles': related to the current 'resemblance fit', but portraits that are more artistically creative or 'artistically fit'. This dual evolving technique of 'patriarchs and strange uncles' attempts to mimic known aspect of creative individuals discussed in Sect. 3, that is the paradoxical technique where creative people use the existence of some strong structural rules (as in the templates of a sonnet, tragedy, or in this case a resemblance to the sitter image) as a resource or base to elaborate new variants beyond that structure (abstracted variation of the sitter image). That is, being novel needs a reference system to rebel and innovate from.

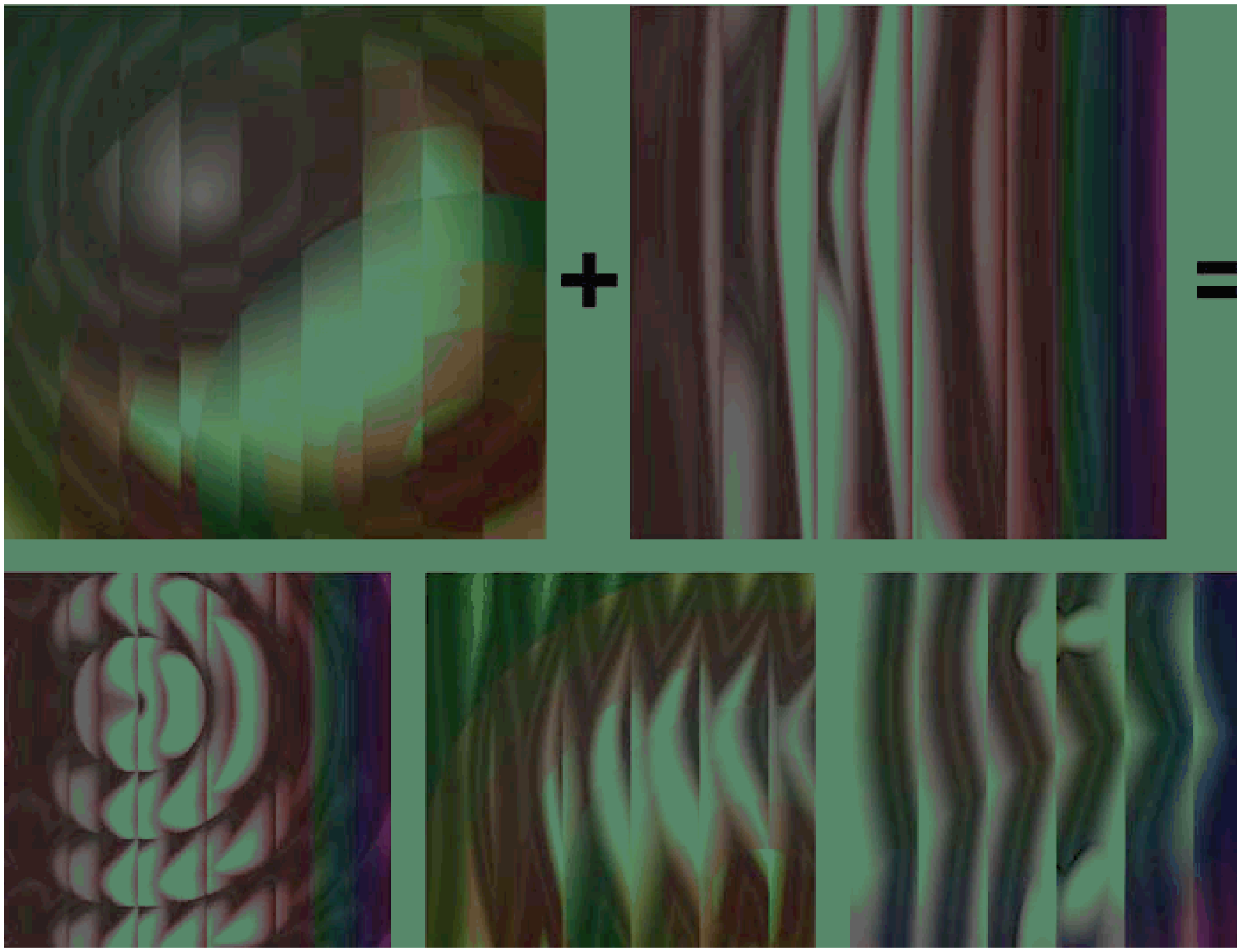

Fig. 3 Two portrait programs are mated together showing merged strategies of the offspring

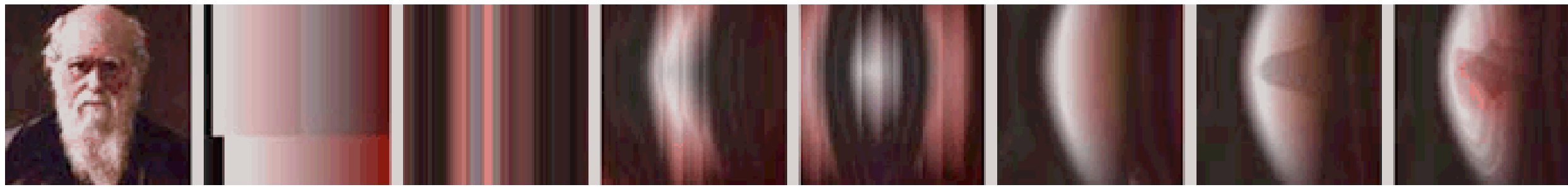

Fig. 4 Source Darwin portrait, part of the fitness function, followed by an evolved progression of portraits of best resemblance

In our system, those individuals that doggedly strive to resemble the Darwin source move the system forward (as they attain the highest resemblance scores and strategically move the system closer to the source image from a resemblance point of view) allowing their related family members to be more innovatively artistic (via large local exploration) as safe variants from the patriarchs. Figure 5 shows both types of individuals working synergistically, while Fig. 4 only contains the 'resemblance' patriarchs. The goal is not to remake the Darwin portrait, but to explore a family tree of related and living portraits that inherit creative painting strategies through an evolutionary process. We aim to eventually iterate a system that can be creative in a range of artistic and design oriented spaces beyond artistic portrait.

## 6 Results

The images in Fig. 5 show selected portraits, going in chronological order. These represent a larger collection, and show both those best at resemblance of the peers, as well as those that are artistically compelling from an abstract portrait perspective. While the overall population improves at resembling Darwin's portrait, what is more interesting to us is the variety of recurring, emergent and merged creative strategies that evolve as the programs in different ways to become better abstract portraitists.

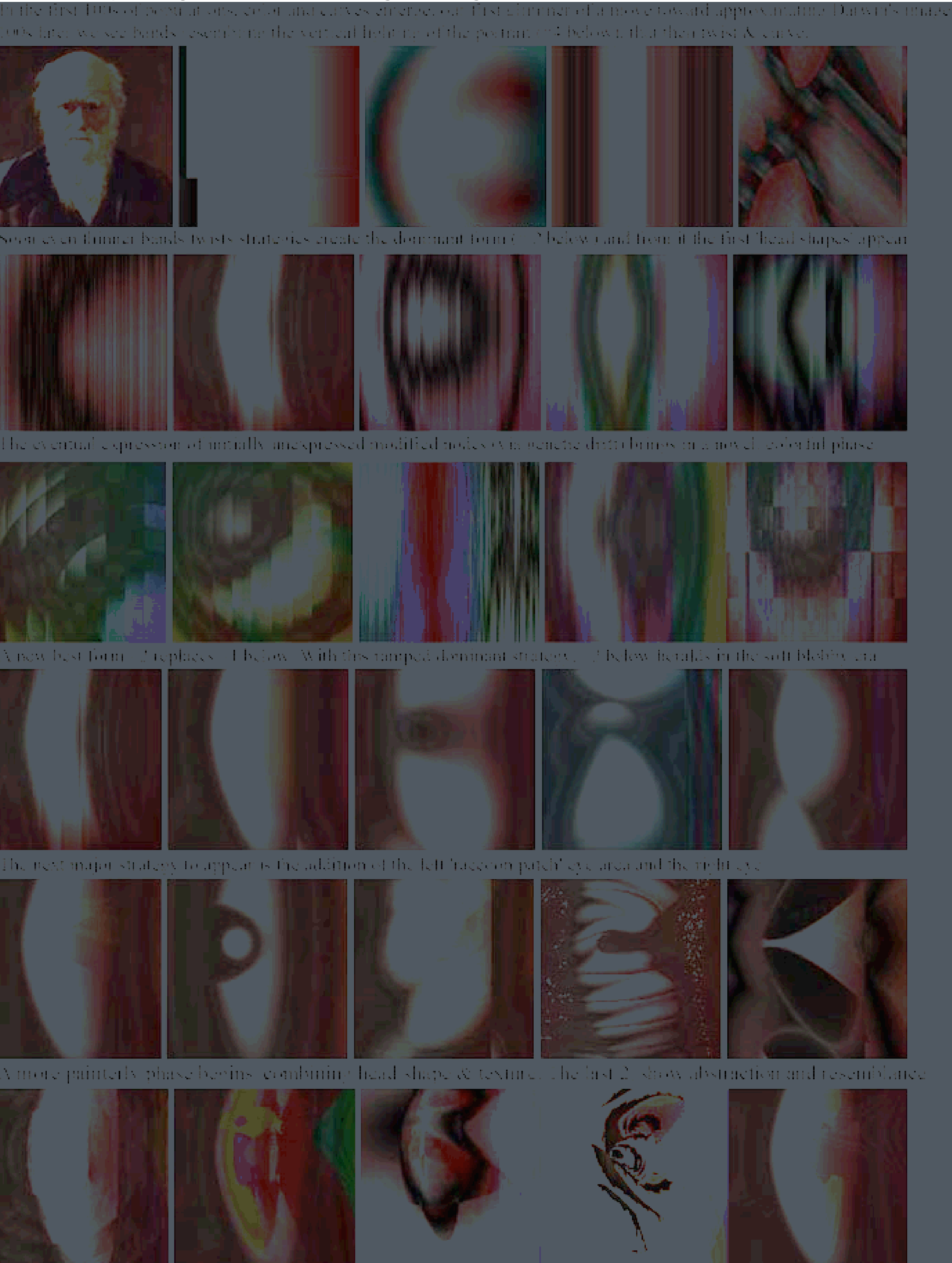

Fig. 5 Portraits in chronological order, selected as examples of the process (from a larger sampling at www.dipaola.org/evolve)

**7 Conclusion and future direction**

We have incorporated research on human creativity into a relativity new form of evolutionary computation, which has been modified to encourage the development of creative, painterly techniques. The domain of portrait painting was chosen because it leans heavily on resemblance (a closed and known issue for computer algorithms), but also has an opened ended creative space and a known portrait sitter/ painter relationship well suited to exploring computer creativity. The program indeed evolves creative strategies to become better abstract portraitists. We are continually refining the painterly portions of the automatic fitness function from lessons learned in past runs, and for our next pass we are adding more creative, structural elements to the current open-ended general system. An ambitious direction for future work is to allow the fitness function to emerge within a neural network or probabilistic cognitive architecture. The rationale for this move is largely theoretical. It has been tentatively proposed that Darwinian algorithms are less than ideal simulations of creative cognition because human creativity is not so much a matter of randomly generating vast quantity of possibilities and selecting the best, but of honing and revising one particular creative product, approximating ever more closely one's insight or vision [12, 13]. To be sure, ideas evolve to some extent when passed on to offspring who, by random accident, put their own spin on them. But to a greater extent they are the product of intuition and strategy in the mind of particular individuals who contemplate them for extended periods. So natural selection is not the agent of change [11].

Moreover even when there is selection going on in a creative thought process, it cannot be described in Darwinian terms because it violates some of the conditions that make natural selection applicable to a formal description of change of state (for example, the fitness function is never the same for all variants; each is evaluated in light of the ones that were generated previously) [13]. The argument here is not that human creativity cannot be formally described as an evolution process; it is that it cannot be described as a selection process. It is proposed that a creative idea evolves through a process of context-driven actualization of potential, or CAP [11; for crossdisciplinary applications of CAP see 14]. That is, an idea exists in a certain state with the potential to change different ways depending on the mind that mulls it over, and the context (situation or environment) that mind encounters. This interaction between mind and context causes an idea to undergo a change of state. Now the idea not only exists in a slightly altered form, but its potential for further change is also altered. This change of state of the idea feeds back on the state of the mind and the state of the context, such that the idea may undergo another change of state, which has associated with it a different set of potential future states. This continues until the interaction between mind and context produces no further change in the idea, at which point the idea is said to be in an eigenstate or end state with respect to this context. Returning to the topic of computer art, our goal is to further the development of the human creativity inspired aspects of the approach initiated here to the point where we can do away with the Darwinian, population-based aspect of the program, replacing it with a CAP approach.

Key to this is increased understanding of how the potentiality of an idea changes and is affected by both the associative structure and the goals and desires of the mind it 'finds itself in'. To this end, future research will involve adding specific painterly and portrait knowledge with the goal of continuing to improve the automatic portrait painter system with human painterly knowledge. To better approximate a human portraitist's technique we are redesigning the functions in the function set to be reactions to the color and position of the sitter image (the current system function set is blind to the sitter image). This way, any decision on a paint stroke output is a direct reaction to the input recognition (what the artist sees in the sitter scene). This would mean that, once a pleasing portrait image/individual is created, the program could use its same painterly strategies on any new sitter image, thereby creating a true portrait painter. A successful portraitist program might even have 'one-man' shows and take commissions, allowing its human creator to play a background role as its talent agent. It could eventually even be bred it with other successful portraitist programs similar to racing horses, allowing for experiments into cultural and collaborative creativity. This 'matching output stroke to input analysis' technique with other modifications would facilitate the realization of another goal: to have resolution-independent portraits, allowing small portrait sizes for speed during the evolving process, but larger sizes that reveal additional painterly and surface details for final artwork—as a human might make many creative sketches before the fully finished work.

We would like to explore the extent to which techniques used here can be transported to other domains such as art and design, music, authoring, HCI, entertainment, and gaming. The mechanisms will be kept general since we believe it is the associative, domain-general (rather than specialized, domain-specific) aspect of a creative architecture (organic or artificial) that is its greatest asset. Finally, we foresee a possible research application as a test bed for simulating creative processes or an educational tool for gaining hands-on understanding of evolutionary and creative processes.

**Acknowledgements**

We would like to thank Laurence Ashmore, Peter Bentley, Julian Miller, and James Walker for their correspondence, as well as Ashmore and Miller for their initial Java-based system that we adapted for our creative experiments. This research partially supported by SSHRC (Gabora) and NSERC (DiPaola) Canada.

1 The variants are said to be 'blind' in the sense that they are generated without foresight as to whether they are a step in the direction of the final creative product.